\documentclass{article}

\usepackage[dblblindworkshop, final]{neurips_2025}

\usepackage[utf8]{inputenc} 
\usepackage[T1]{fontenc}    
\usepackage{hyperref}       
\usepackage{url}            
\usepackage{booktabs}       
\usepackage{amsfonts}       
\usepackage{nicefrac}       
\usepackage{microtype}      
\usepackage{xcolor}         

\newcommand{\DefinedAs}[0]{\mathrel{\mathop:}=}
\usepackage{wrapfig}
\usepackage{bbm}
\def\Real{\mathbb{R}}
\usepackage{color}

\usepackage{amsmath,amssymb}
\usepackage{amsthm,bm}
\usepackage{cases}
\usepackage[normalem]{ulem}
\usepackage{multirow}

\usepackage{amsmath}   
\usepackage{mathtools} 

\DeclarePairedDelimiter{\norm}{\lVert}{\rVert}

\newtheorem{theorem}{Theorem}
\newtheorem{assumption}{Assumption}

\newtheorem{lemma}{Lemma}

\newenvironment{myproof}{{\noindent\it Proof. }}{\hfill $\square$ \par}

\usepackage{caption}
\usepackage{dsfont}

\usepackage{comment}

\usepackage{subfigure}
\usepackage{graphicx}

\usepackage{algorithm, algpseudocode}
\usepackage{float}

\input{mysymbol.sty}

\usepackage{ulem}

\usepackage[most]{tcolorbox} 
\newtcolorbox{promptbox}[1]{
  enhanced,
  colback=red!5!gray!10,
  colframe=red!50!black,
  arc=2mm,
  boxrule=1pt,
  title=#1,
  fonttitle=\bfseries\color{white},
  coltitle=red!50!black,
  breakable=false,
  width=\textwidth
}

\workshoptitle{Efficient Reasoning}

\title{ConstrainedSQL: Training LLMs for Text2SQL via Constrained Reinforcement Learning}

\author{%
  Weiqin Chen$^{\spadesuit, \clubsuit}$ \quad Nhan Huu Pham$^\clubsuit$ \quad Michael Robert Glass$^\clubsuit$ \quad Long Hai Vu$^\clubsuit$ \\ \textbf{Gaetano Rossiello}$^\clubsuit$ \quad \textbf{Dharmashankar Subramanian}$^\clubsuit$ \quad \textbf{Santiago Paternain}$^\spadesuit$  \\ \\
  $^\spadesuit$ Rensselaer Polytechnic Institute \\
  $^\clubsuit$ IBM Research
}

\begin{document}

\maketitle

\begin{abstract}
  Reinforcement learning (RL) has demonstrated significant promise in enhancing the reasoning capabilities of Text2SQL LLMs, especially with advanced algorithms such as GRPO and DAPO. However, the performance of these methods is highly sensitive to the design of reward functions. Inappropriate rewards can lead to ``reward hacking'', where models exploit loopholes in the reward structure to achieve high scores without genuinely solving the task. This work considers a constrained RL framework for Text2SQL that incorporates natural and interpretable reward and constraint signals, while dynamically balancing trade-offs among them during the training. We establish the theoretical guarantees of our constrained RL framework and our numerical experiments on the well-known Text2SQL datasets substantiate the improvement of our approach over the state-of-the-art RL-trained LLMs.
\end{abstract}

\section{Introduction}

Text2SQL aims to convert natural language questions to structured SQL statements, thus enabling the interactions with the database without any expertise and bridging the gap between the human language and database queries~\cite{liu2024survey}. Although large language models (LLMs) with supervised fine-tuning (SFT) have shown great progress, they are often limited to the database schema structure and data scale.
Recently, the success of large reasoning models such as DeepSeek-R1~\cite{guo2025deepseek} and OpenAI o1~\cite{jaech2024openai} has demonstrated the potential of reinforcement learning (RL) in training the reasoning capabilities of LLMs. In particular, RL enables dynamic adjustment of strategies through interaction with the environment, a capability that is crucial for complex reasoning tasks~\cite{guo2025deepseek}.
However, despite the advances in RL algorithms e.g., GRPO~\cite{shao2024deepseekmath} and DAPO~\cite{yu2025dapo}, the policies are highly sensitive to the reward design. This is not surprising, since the best imaginable algorithm can only perform as well as the task it is asked to solve (the reward it must maximize).
In particular, current state-of-the-art (SOTA) RL-trained Text2SQL LLMs such as SQL-R1~\cite{ma2025sql} and Reasoning-SQL~\cite{pourreza2025reasoning} rely on manually designed rewards, which may be neither efficient nor reliable. Notice that inappropriate rewards can lead to ``reward hacking'', where models exploit loopholes in the reward structure to achieve high scores without genuinely solving the task. Moreover, carefully and manually designed rewards can be costly and often generalize poorly to new tasks.

In this context, we propose ConstrainedSQL that trains Text2SQL LLMs via constrained RL~\cite{paternain2019constrained, chen2024probabilistic}. In particular, we design several natural and interpretable reward and constraint metrics for Text2SQL, and dynamically balance the trade-off among them during the RL training. Furthermore, we establish the theoretical guarantee of our method in terms of a bounded primal-dual gap that is dominated by a parameterization gap. We conduct extensive experiments on standard Text2SQL benchmarks including Spider (Dev), Spider (Test), and BIRD (Dev), comparing our approach with both SOTA RL-trained and SFT-based Text2SQL models. The results demonstrate that our method outperforms SOTA RL-trained models and achieves performance comparable to SFT-based models, despite the latter using roughly $50\times$ more training samples.

\section{ConstrainedSQL}

\subsection{Dataset}
In this work, we extract our dataset from SynSQL-2.5M~\cite{li2025omnisql}, which is the first million-scale synthetic Text2SQL dataset and contains more than $2.5$ million high-quality samples. SFT often demands a substantial amount of samples, which is time consuming and resource demanding and is typically limited to producing only simple to moderate SQL queries. This occurs because SFT often tends to memorize only the samples encountered in its training dataset~\cite{chu2025sft}.
To explore the capability of LLM to generate complex SQL queries under limited samples, this work considers a dataset of $(x_i, y_i^\star)_{i=1}^{5000}$, where $x_i$ denotes the prompt (encompass the natural language question) and $y_i^\star$ its corresponding ground-truth response (ground-truth SQL). Notably, these $5000$ samples are randomly drawn from SynSQL-2.5M among those that are deemed \emph{Complex} (other categories in the dataset are \textit{Simple}, \textit{Moderate}, etc.). Having established the dataset, our work focuses on fine-tuning the Text2SQL LLM using RL rather than SFT, as RL enables dynamic adjustment of the strategy through interactions with the environment and can offer superior performance in complex reasoning tasks. The procedure of RL training is introduced in the following subsection.

\subsection{Training of ConstrainedSQL}

Denote by $\ccalX$ and $\ccalY$ the prompt space and the LLM response space. An LLM policy $\pi_\theta : \ccalX \to \Delta(\ccalY)$ represents a mapping from the prompt space to the set of all distributions over the response space $\Delta(\ccalY)$, where $\theta$ denotes the parameters in the LLM. The reward $r: \ccalX \times \ccalY \to \Real$ evaluates the quality of the response generated by the LLM policy, which can be instantiated either as a reward model trained on human preference datasets or as a human-designed reward function. Consider a reference LLM policy $\pi_{\text{ref}}$ (e.g., a SFT model). The training of the LLM policy via RL is formulated as
\begin{align}\label{eqn_unconstrained_RL}
        \max_{\theta\,\in\,\Theta}
        \; & \;
        \mathbb{E}_{x \,\sim\, \mathcal{D}}
       \left[\, 
       \mathbb{E}_{y  \,\sim \,\pi_\theta}[\,r_0(x, y)\,]
       -
       \beta\,
       D_{\text{KL}}(\pi_{\theta}(\cdot\,\vert\, x) \,\Vert\, \pi_{\text{ref}}(\cdot \, \vert \, x))
       \,\right],
\end{align}
where $ D_{\text{KL}}(\cdot||\cdot)$ denotes the KL-divergence and  $\beta$ represents the regularization constant. Then, the objective of RL is to optimize the LLM policy parameters to maximize the expected reward while ensuring the LLM policy does not deviate excessively from the reference LLM. 
In the context of Text2SQL, the reward structure often adopts a composite reward with several sub-rewards such as format reward, execution reward, etc \cite{ma2025sql}. To this end, manually designing each reward and carefully tuning the corresponding weights between them becomes highly challenging. Moreover, this can lead to reward hacking. To address this issue, we adopt a constrained RL framework for training the LLM
\begin{align}\label{eqn_constrained_RL_parameterize}
        P^\star = \max_{\theta\,\in\,\Theta}
        \; & \;
        \mathbb{E}_{x \,\sim\, \mathcal{D}}
       \left[\, 
       \mathbb{E}_{y  \,\sim \,\pi_\theta}[\,r(x, y)\,]
       -
       \beta\,
       D_{\text{KL}}(\pi_{\theta}(\cdot\,\vert\, x) \,\Vert\, \pi_{\text{ref}}(\cdot \, \vert \, x))
       \,\right]\
       \nonumber \\
       \text{s.t.} \; &\;
       \mathbb{E}_{x \,\sim\, \mathcal{D}}
       \left[\, 
       \mathbb{E}_{y  \,\sim \,\pi_\theta}[\, c_i(x, y)]\,]
       \,\right] 
       \;\geq\;
       b_i,\; \, i = 1,\ldots, m
\end{align}
where we consider a single reward function $r$, and $c_i$ and $b_i$ denote the $i$-th constraint and the corresponding threshold. The reward $r$ and constraints $c_i$ should be designed to align with the problem’s objective and requirements. For brevity, let us define $g_i(x, y) = c_i(x, y) - b_i$.
Section~\ref{sec_experiments} presents a specific instantiation of problem~\eqref{eqn_constrained_RL_parameterize} in the Text2SQL setting, where we consider natural and interpretable reward and constraints without the need for carefully designing each component. Before proceeding, we introduce in the next section the theoretical guarantees of problem~\eqref{eqn_constrained_RL_parameterize}.

\section{Theoretical Guarantees}

A common approach to solve the constrained optimization problem~\eqref{eqn_constrained_RL_parameterize} is to solve its unconstrained approximation using the Lagrangian
\begin{equation}\label{eqn_lagrangian_parametrized}
     L(\pi_\theta, \lambda) :=
    \mathbb{E}_{x \sim \ccalD } 
\left[\,
 \mathbb{E}_{y \,\sim\, \pi_{\theta}}\left[ \, r(x, y)
 + 
\lambda^\top g(x, y)
 \,\right]
 -
\beta\,
D_{\mathrm{KL}}(\pi_{\theta}(\cdot \mid x) \, \| \, \pi_{\text{ref}}(\cdot \mid x)) 
\,\right],
\end{equation}
where $g=[g_1, g_2, \ldots, g_m]^\top$ and $\lambda=[\lambda_1, \lambda_2, \ldots, \lambda_m]^\top$ denote the constraints and the corresponding Lagrange multipliers that trade-off the objective and the corresponding constraints. Indeed, for large values of $\lambda_i$ solutions will prioritize the $i$-th constraint, whereas for small values of $\lambda_i$ the solutions will focus on the objective and other constraints. To this end, each constraint threshold $b_i$ will correspond to a specific optimal Lagrange multiplier $\lambda_i^\star$. The dual function associated with \eqref{eqn_lagrangian_parametrized} is defined as $D(\lambda) \DefinedAs \max_{\theta\,\in\,\Theta} L(\pi_\theta, \lambda)$. Subsequently, the dual problem for \eqref{eqn_constrained_RL_parameterize} is given by
\begin{equation}\label{eqn_dual_function_parametrized}
    D^\star = \min_{\lambda \,\geq\, 0}\; D(\lambda).
\end{equation}
It is worth highlighting that problem \eqref{eqn_dual_function_parametrized} is always convex despite the non-convexity of problem \eqref{eqn_constrained_RL_parameterize}~\cite{boyd2004convex}, thus enabling the employment of gradient-based approaches. 
%
%
However, the primal and dual problems become equivalent solely in certain scenarios such as convex problems, as illustrated below
\begin{align}\label{eqn_constrained_RL_policy}
        \max_{\pi\,\in\,\Pi}
        \; & \;
        \mathbb{E}_{x \,\sim\, \mathcal{D}}
       \left[\, 
       \mathbb{E}_{y  \,\sim \,\pi}[\,r(x, y)\,]
       -
       \beta\,
       D_{\text{KL}}(\pi(\cdot\,\vert\, x) \,\Vert\, \pi_{\text{ref}}(\cdot \, \vert \, x))
       \,\right]\
       \nonumber \\
       \text{s.t.} \; &\;
       \mathbb{E}_{x \,\sim\, \mathcal{D}}
       \left[\, 
       \mathbb{E}_{y  \,\sim \,\pi}[\,g_i(x, y)]\,]
       \,\right] 
       \;\geq\;
       0,\; \, i = 1,\ldots, m.
\end{align}
The previous problem is convex as the expectation is linear and the KL divergence is convex in the distribution $\pi(\cdot \mid x)$ over responses. Yet, \eqref{eqn_constrained_RL_parameterize} is generally a non-convex optimization problem, primarily due to factors such as the nonlinearities introduced by the Transformer blocks~\cite{vaswani2017attention} in the LLM policy.
Thus, it is significant to investigate the  primal-dual gap between problems \eqref{eqn_constrained_RL_parameterize} and \eqref{eqn_dual_function_parametrized}. To proceed, we rely on the following assumptions.
\begin{assumption}[Boundness]\label{ass_boundness}
    There exists a constant $B$ such that, for any $(x, y) \in \ccalX \times \ccalY$ and $i \in \{1, 2, \ldots, m\}$, $\max \left(|r(x, y)|, |g_i(x, y)| \right) \leq B$.
\end{assumption}
\begin{assumption}[Parametrization Gap]\label{ass_parameter_gap}
    There exists a constant $\nu$ such that for any policy $\pi\in \Pi$ and any $x\in\mathcal{X}$, there exists $\theta\in\Theta$ such that 
    $\max \left( ||\pi_\theta(\cdot\,\vert\,x) - \pi(\cdot\,\vert\,x)||_1, |D_{\text{\normalfont KL}}(\pi_\theta(\cdot\,\vert\, x) \,\Vert\, \pi_{\text{\normalfont ref}}(\cdot \, \vert \, x))- D_{\text{\normalfont KL}}(\pi(\cdot\,\vert\, x) \,\Vert\, \pi_{\text{\normalfont ref}}(\cdot \, \vert \, x))| \right) \leq \nu$.
\end{assumption}
\begin{assumption}[Strict Feasibility]\label{ass_feasibility}
    There exists $\theta \in \Theta$ and $\xi>0$ such that $\mathbb{E}_{x\sim \ccalD}
        \left[\, 
        \mathbb{E}_{y  \,\sim \,\pi_{\theta}(\cdot \,\vert\, x)}[\,g_i(x, y) \,\right] \geq B\nu + \xi \,$ for all $i=1,\ldots,m$.
\end{assumption}

Assumption~\ref{ass_boundness} is common in the literature \cite{azar2017minimax, wei2020model, zhang2021near, chen2024domain}. In
particular, in the case of finite state-action spaces such as language and SQL query space, it is always possible to design the reward to avoid the possibility of being unbounded. Assumption~\ref{ass_parameter_gap} states that the parameterization space $\Theta$ in \eqref{eqn_constrained_RL_parameterize} covers the policy space $\Pi$ in \eqref{eqn_constrained_RL_policy} up to an error $\nu$.
Assumption~\ref{ass_feasibility} is also a standard assumption in the primal-dual theory~\cite{boyd2004convex}, positing the existence of a strictly feasible parameter.

\begin{theorem}[Primal-dual gap]\label{theorem_primal_dual_gap}
    Let Assumptions~\ref{ass_boundness}, \ref{ass_parameter_gap} and~\ref{ass_feasibility} hold. Denote by $\tilde{D}(\lambda)$ the dual function of \eqref{eqn_constrained_RL_policy} and define $\lambda_\nu^\star = \argmin_{\lambda\,\geq\;0} \tilde{D}(\lambda) - B\nu\norm{\lambda}_1$. Then, it holds that
    \begin{equation}\label{eqn_primal_dual_gap}
    0
    \; \leq \;
    D^\star  
    -
    P^\star
    \; \leq \;
     \left(\beta+B+B\norm{\lambda_\nu^\star}_1 \right)\nu.
    \end{equation}
\end{theorem}
\begin{myproof}
    Refer to Appendix~\ref{append_theorem_primal_dual_gap}.
\end{myproof}
Notably, Theorem~\ref{theorem_primal_dual_gap} implies that the primal-dual gap between \eqref{eqn_constrained_RL_parameterize} and \eqref{eqn_dual_function_parametrized} is dominated by the parameterization gap $\nu$. In fact, a zero parameterization gap guarantees a zero primal-dual gap despite the non-convexity of \eqref{eqn_constrained_RL_parameterize}~\cite{paternain2019constrained,zhang2025alignment}.

\begin{algorithm}[t]
  \caption{Primal Dual Algorithm}
  \label{alg_pd} 
\begin{algorithmic}[1]
 \renewcommand{\algorithmicrequire}{\textbf{Input:}}
 \renewcommand{\algorithmicensure}{\textbf{Let}}
 \Require Initial Lagrange multipliers $\lambda^0=[\lambda_1^0, \lambda_2^0, \ldots, \lambda_m^0]^\top$, constraint thresholds $b=[b_1, b_2, \ldots, b_m]^\top$, primal learning rate $\eta_\theta$, dual learning rate $\eta_\lambda $, reference model $\pi_{\text{ref}}$
 \Ensure $\pi_{\theta^0} = \pi_{\text{ref}}$
  \For {$k = 0,1,\ldots $}
  \State Update the primal variable (LLM policy parameters) using GRPO
  $$
  \theta^{k+1} = \theta^k+ \eta_{\theta} \nabla_\theta^{\text{GRPO}} L(\pi_{\theta^k},\lambda^k)
   $$
   \State Update the dual variable (Lagrange multipliers) using dual gradient descent
  $$
  \lambda^{k+1} = \left[\lambda^k- \eta_{\lambda} \, \left( c_{\pi_{\theta^{k+1}}} - b \right) \right]_+
    $$
  \EndFor
 \end{algorithmic}
 \end{algorithm}

\section{Experiments}
\label{sec_experiments}

In this section, we demonstrate the numerical performance of our ConstrainedSQL method. Our implementation (see Algorithm~\ref{alg_pd}) is built on the well-known VERL library~\cite{sheng2025hybridflow} and the GRPO algorithm~\cite{shao2024deepseekmath}. The example of the prompt and the response of our ConstrainedSQL is provided in Appendix~\ref{append_prompt_response}, where the prompt includes the natural language question and the response encompasses a reasoning process within the \texttt{<think>...</think>} tag and the answer and corresponding SQL within the  \texttt{<answer>...</answer>} tag.

\textbf{Reward/Constraints.} In this work, we adopt natural and interpretable reward and constraint designs for problem~\eqref{eqn_constrained_RL_parameterize} with each represented by an indicator function. 
Specifically, $r=1$ if the generated SQL yields the correct query result (matching the ground-truth) and $r=0$ otherwise; $g_\text{format} =1$ if the response from the LLM policy shows the correct format (see the example of response format in Appendix~\ref{append_prompt_response}) and  $g_\text{format} =0$ otherwise; $g_\text{execution} =1$ if the generated SQL is executable and $g_\text{execution} =0$ otherwise; $g_\text{length} =1$ if the response is longer than a pre-designed threshold ($300$ in this work) and $g_\text{execution} =0$ otherwise; $g_\text{answer} =1$ if the proportion of the answer in the generated response is within a pre-designed threshold ($25\%-75\%$ in this work) and $g_\text{answer} =0$ otherwise; $g_\text{SQL} =1$ if the generated SQL is longer than a pre-designed threshold ($25\%-$ in this work) of the answer and $g_\text{SQL} =0$ otherwise. All corresponding constraint thresholds are set to be $95\%$. Then with the indicator-function designs for the reward and constraints, problem \eqref{eqn_constrained_RL_parameterize} aims to maximize the probability of generated SQL matching the ground-truth while each of the constraints is satisfied (e.g., the response format being correct) with probabilities at least $95\%$.

\begin{table}[h]
\centering
\caption{Performance comparison (Execution Accuracy) of 3B models}
\renewcommand{\arraystretch}{1.25} %
\label{tab_3B}
\begin{tabular}{|cccc|}
\hline
\textbf{Dataset} & \textbf{\begin{tabular}[c]{@{}c@{}}SQL-R1 (reproduced)\end{tabular}} & \textbf{\begin{tabular}[c]{@{}c@{}}SQL-R1 (reported) \end{tabular}} & \textbf{ConstrainedSQL} \cr
\hline

Spider (Dev) & $79.3$ & $78.1$ & $\textbf{82.0}$  \cr

Spider (Test) & $79.7$ & $78.9$ & $\textbf{81.3}$ \cr

BIRD (Dev) & $51.3$ & $54.6$ & $\textbf{55.0}$  \cr

\hline
\end{tabular}
\end{table}

\textbf{Baselines.} We consider SQL-R1~\cite{ma2025sql} as our main baseline as it is the SOTA RL-based Text2SQL approach. In addition, we compare with other SFT-based methods such as OmniSQL~\cite{li2025omnisql} that often consumes a larger number of samples. To ensure a fair comparison, all methods are built on Qwen2.5-Coder series models~\cite{hui2024qwen2}.

\textbf{Benchmarks.} The ConstrainedSQL, along with all other baselines, is evaluated on two widely-used Text2SQL benchmarks: Spider~\cite{yu2018spider} and BIRD~\cite{li2023can}. The Spider dataset consists of 10,181 natural language questions paired with 5,693 complex SQL queries across 200 databases and 138 domains. The BIRD dataset contains 12,751 Text2SQL pairs covering 95 databases across 37 specialized domains.
We adopt the standard Execution Accuracy metric for both Spider and BIRD benchmarks, which measures the proportion of questions that yield outcomes consistent with their corresponding ground-truth queries across all query requests.

\begin{table}[h]
\centering
\caption{Performance comparison (Execution Accuracy) of 7B models}
\renewcommand{\arraystretch}{1.25} %
\label{tab_7B}
\begin{tabular}{|cccc|}
\hline
\textbf{Dataset} & \textbf{\begin{tabular}[c]{@{}c@{}}SQL-R1 (reproduced)\end{tabular}} & \textbf{\begin{tabular}[c]{@{}c@{}}SQL-R1 (reported)\end{tabular}} & \textbf{ConstrainedSQL} \cr
\hline

Spider (Dev) & $83.3$ & $84.5$ & $\textbf{85.2}$  \cr

Spider (Test) & $83.2$ & $86.1$ & $\textbf{86.7}$  \cr

BIRD (Dev) & $60.9$ & $\textbf{63.1}$ & $61.9$  \cr

\hline
\end{tabular}
\end{table}

\begin{table}[h]
    \caption{Execution Accuracy~(\%) of different Text2SQL methods on Spider and BIRD benchmark.}
    \centering
    \begin{tabular}{ccccc}
    \toprule
    \textbf{Text2SQL Method} &
      \textbf{Base Model} &
      \textbf{\begin{tabular}[c]{@{}c@{}}Spider\\ (Dev)\end{tabular}} &
      \textbf{\begin{tabular}[c]{@{}c@{}}Spider \\ (Test)\end{tabular}} &
      \textbf{\begin{tabular}[c]{@{}c@{}}BIRD\\ (Dev)\end{tabular}}  \\
      \midrule
        Alpha-SQL~\cite{matsui2022alphasql} & Qwen2.5-Coder-7B &  84.0 & - & 66.8 \\
        SQL-o1~\cite{lyu2025sql} & Qwen2.5-Coder-7B & 84.7 & 85.1 & 66.7 \\
        OmniSQL~\cite{li2025omnisql} & Qwen2.5-Coder-7B & 85.5 & 88.9 & 66.1 \\
        DeepRetrieval~\cite{jiang2025deepretrieval}  & Qwen2.5-Coder-7B & - & 76.1 & 56.0 \\
        SQL-R1 & Qwen2.5-Coder-7B & 84.5 & 86.1 & 63.1 \\ 
    \midrule
    \textbf{ConstrainedSQL (Ours)} & \textbf{Qwen2.5-Coder-7B} & \textbf{85.2} & \textbf{86.7} & \textbf{61.9} \\
    \bottomrule
    \end{tabular}
    \label{tab_other_sft_method}
\end{table}

\textbf{Results.} Our main results are presented in Tables~\ref{tab_3B} and \ref{tab_7B}. We compare our ConstrainedSQL approach with two versions of SQL-R1 method: reproduced (our reproduced results using the official implementation of SQL-R1) and reported (the results reported in the original SQL-R1 paper). Table~\ref{tab_3B} summarizes the experimental results for the 3B models, showing that our ConstrainedSQL method consistently outperforms the two SQL-R1 baselines across all benchmarks. The results for the 7B models are reported in Table~\ref{tab_7B}. ConstrainedSQL continues to outperform both SQL-R1 baselines on the Spider (Dev) and Spider (Test) datasets. On the BIRD (Dev) dataset, ConstrainedSQL surpasses the SQL-R1 (reproduced) baseline and delivers performance comparable to SQL-R1 (reported).
Meanwhile, it is worth noting that the SQL-R1 baselines rely on carefully crafted reward designs~\cite{ma2025sql}, whereas our approach simply adopts the natural and interpretable reward and constraint designs.

\begin{figure}[ht]
\centering 
\setcounter{subfigure}{0}
\subfigure[Reward]
{
	\begin{minipage}{0.31\linewidth}
	\centering 
	\includegraphics[width=1.0\columnwidth]{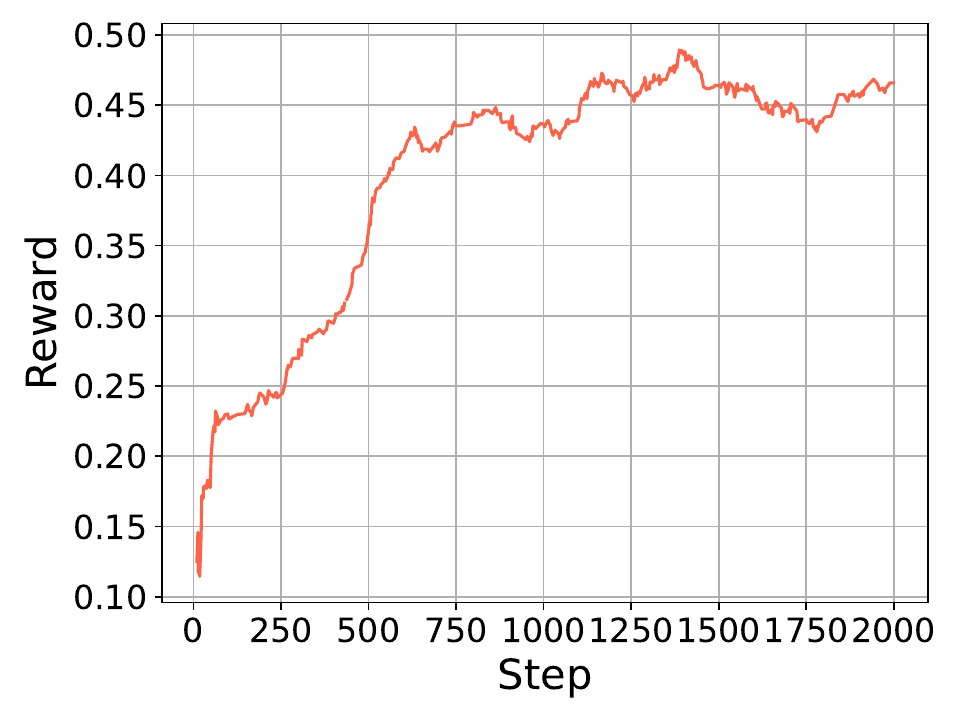}
	\end{minipage}
}
\subfigure[Format Constraint]
{
	\begin{minipage}{0.31\linewidth}
	\centering 
	\includegraphics[width=1.0\columnwidth]{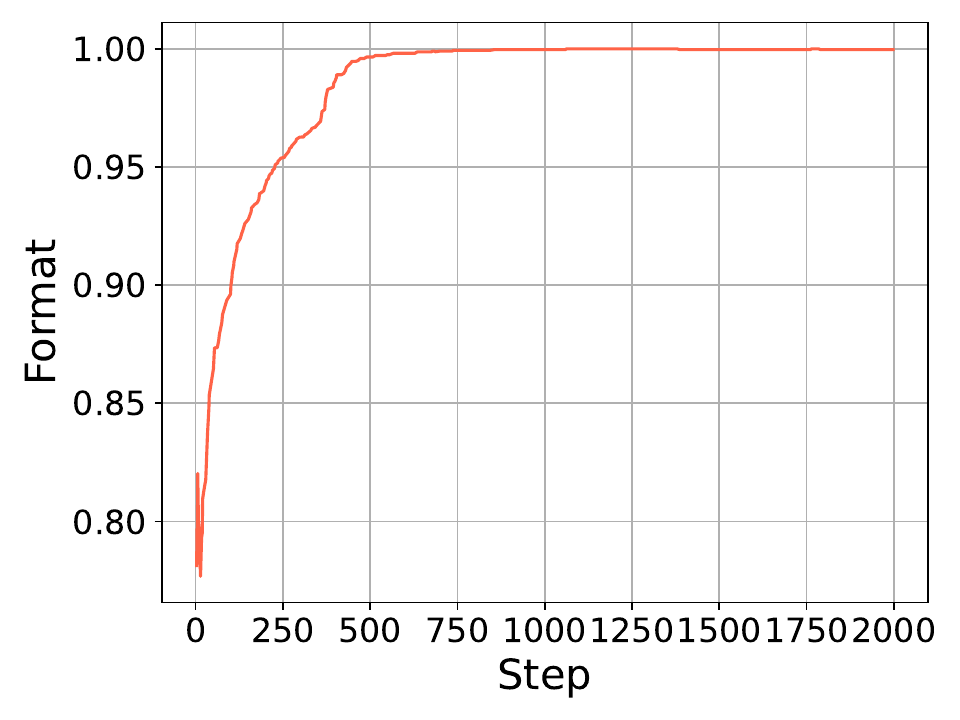}
	\end{minipage}
}
\subfigure[Execution Constraint]
{
	\begin{minipage}{0.31\linewidth}
	\centering 
	\includegraphics[width=1.0\columnwidth]{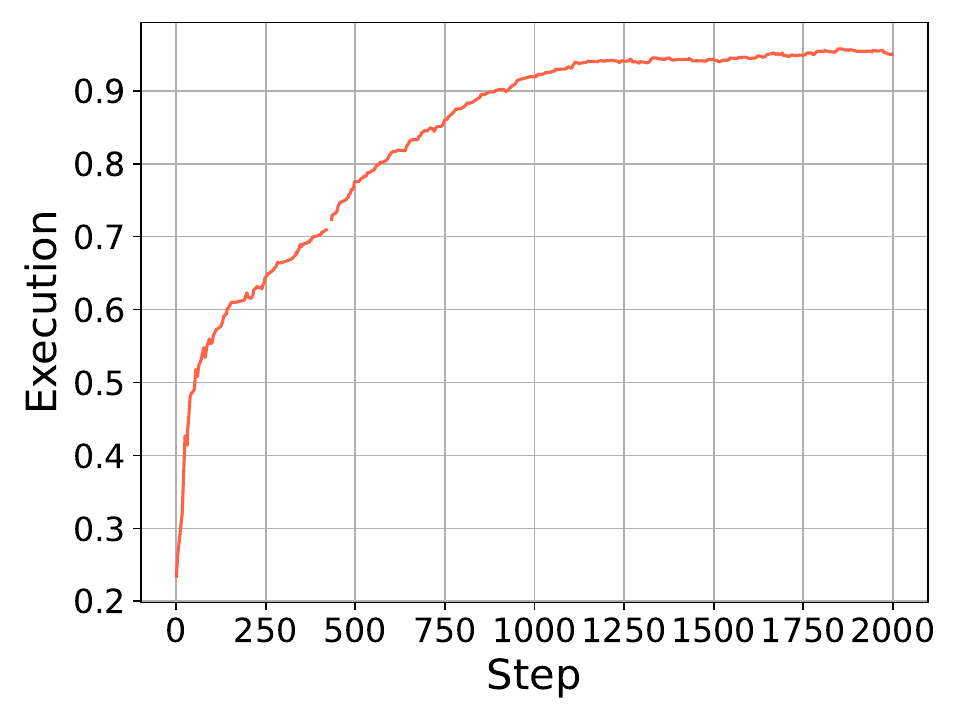}
	\end{minipage}
}
\\
\subfigure[Length Constraint]
{
	\begin{minipage}{0.31\linewidth}
	\centering 
	\includegraphics[width=1.0\columnwidth]{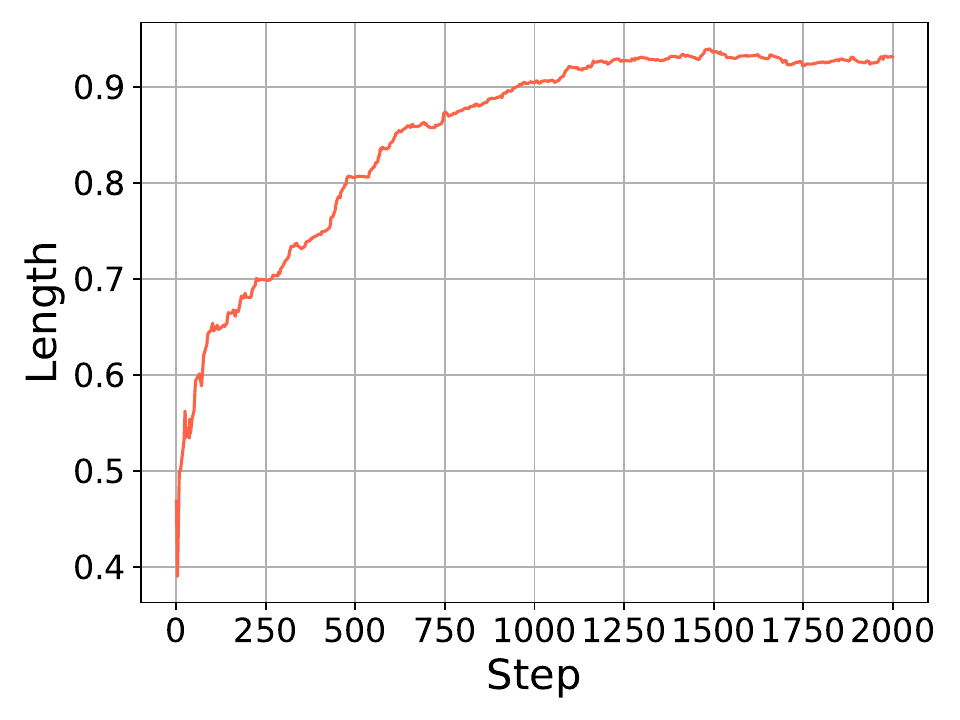}
	\end{minipage}
}
\subfigure[Answer Constraint]
{
	\begin{minipage}{0.31\linewidth}
	\centering 
	\includegraphics[width=1.0\columnwidth]{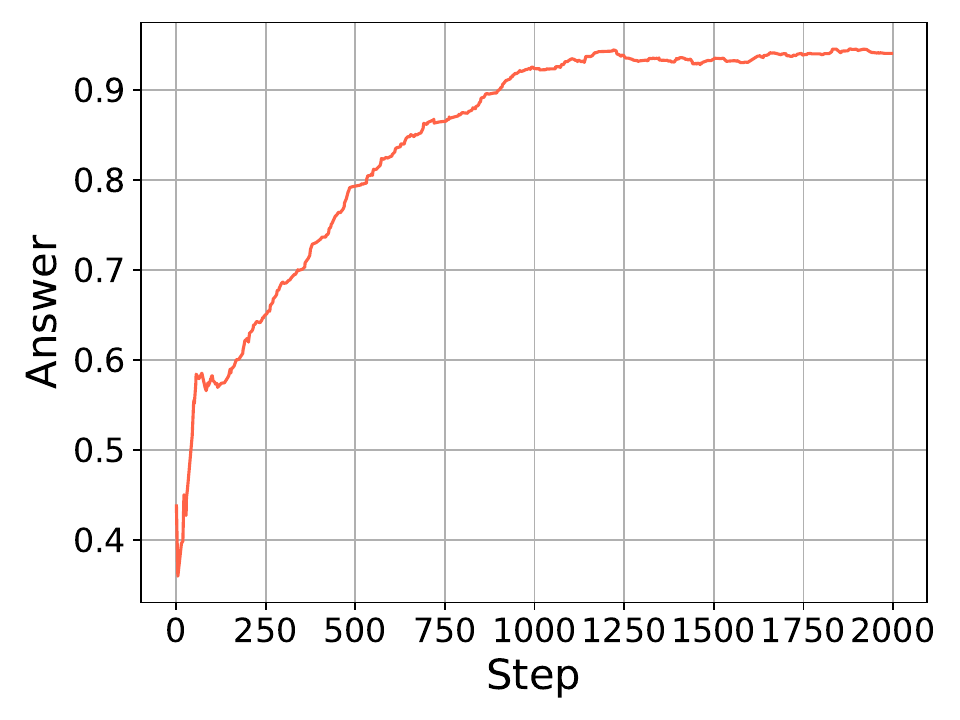}
	\end{minipage}
}
\subfigure[SQL Constraint]
{
	\begin{minipage}{0.31\linewidth}
	\centering 
	\includegraphics[width=1.0\columnwidth]{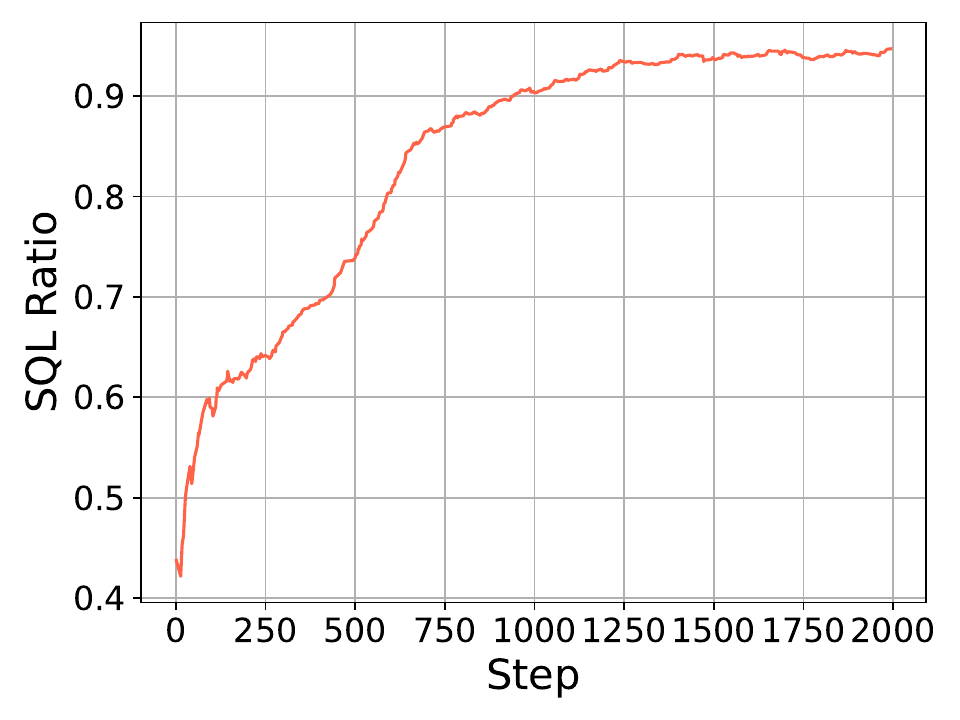}
	\end{minipage}
}
\caption{Training curves of the reward and constraints in ConstrainedSQL.}
\label{fig_train_curve_7B}
\end{figure}

In Table~\ref{tab_other_sft_method}, we also compare ConstrainedSQL with other SFT-based Text2SQL methods such as OmniSQL~\cite{li2025omnisql}. Notably, ConstrainedSQL achieves comparable performance to these SFT-based methods, despite their use of up to 2.5 million training samples, while we rely on only 5,000 samples.
Figure~\ref{fig_train_curve_7B} depicts the evolution of each metric (both reward and constraints) during the training of ConstrainedSQL, demonstrating substantial improvements and clear convergence for all metrics.

\section{Conclusion}
In this work, we introduce ConstrainedSQL that leverages natural and interpretable reward and constraint signals for training Text2SQL LLMs. By dynamically balancing trade-offs among multiple metrics, our approach tackles the issues such as reward hacking. We provide theoretical guarantees for our framework and demonstrate through extensive experiments on the standard Text2SQL benchmarks (Spider and BIRD) that it outperforms the current SOTA RL-trained LLMs. These results highlight the potential of constrained RL to improve both the reliability and interpretability of LLMs' post-training.

\section*{Acknowledgments}
This work was supported by IBM through the IBM-Rensselaer Future of Computing Research Collaboration.

\clearpage

\bibliography{neurips_2025}
\bibliographystyle{unsrt}

\clearpage


\appendix

\section{Proof of Theorem~\ref{theorem_primal_dual_gap}}
\label{append_theorem_primal_dual_gap}
To prove Theorem~\ref{theorem_primal_dual_gap}, we rely on the following technical lemma.
\begin{lemma}\label{lemma_append_bound_B_nu}
    Let Assumptions~\ref{ass_boundness} and \ref{ass_parameter_gap} hold. For any policy $\pi\in \Pi$, there exists $\theta\in\Theta$ such that the following inequalities hold
    \begin{align}
    &\left| \mathbb{E}_{x \,\sim\, \mathcal{D}}
       \left[\, 
       \mathbb{E}_{y  \,\sim \,\pi_{\theta}}[\,r(x, y)]
       \,\right] - \mathbb{E}_{x \,\sim\, \mathcal{D}}
       \left[\, 
       \mathbb{E}_{y  \,\sim \,\pi}[\,r(x, y)]
       \,\right]  \right|  \leq B \nu, \label{eqn_append_lemma_reward} \\
    &\left| \mathbb{E}_{x \,\sim\, \mathcal{D}}
       \left[\, 
       \mathbb{E}_{y  \,\sim \,\pi_{\theta}}[\,g_i(x, y)]
       \,\right] - \mathbb{E}_{x \,\sim\, \mathcal{D}}
       \left[\, 
       \mathbb{E}_{y  \,\sim \,\pi}[\,g_i(x, y)]
       \,\right]  \right|  \leq B \nu, \; \, i = 1,\ldots, m \label{eqn_append_lemma_constraint}
\end{align}
\end{lemma}

\begin{myproof}
We start by proving \eqref{eqn_append_lemma_reward}. For any policy $\pi\in \Pi$, there exists $\theta\in\Theta$ such that
\begin{align}
    &\left| \mathbb{E}_{x \,\sim\, \mathcal{D}}
       \left[\, 
       \mathbb{E}_{y  \,\sim \,\pi_{\theta}}[\,r(x, y)]
       \,\right] - \mathbb{E}_{x \,\sim\, \mathcal{D}}
       \left[\, 
       \mathbb{E}_{y  \,\sim \,\pi}[\,r(x, y)]
       \,\right]  \right| \\
       &=\left| \mathbb{E}_{x \,\sim\, \mathcal{D}}
       \left[\, 
       \mathbb{E}_{y  \,\sim \,\pi_{\theta}}[\,r(x, y)] - \mathbb{E}_{y  \,\sim \,\pi}[\,r(x, y)]
       \,\right]  \right| \\
       &=\left| \mathbb{E}_{x \,\sim\, \mathcal{D}}
       \left[\, \sum_{y \in \ccalY} \pi_{\theta} (y \mid x) r(x, y) - \sum_{y \in \ccalY} \pi (y \mid x) r(x, y) \right]  \right| \\
       &=\left| \mathbb{E}_{x \,\sim\, \mathcal{D}}
       \left[\, \sum_{y \in \ccalY} \left(\pi_{\theta} (y \mid x) - \pi (y \mid x) \right) r(x, y) \right]  \right|.
\end{align}
Given that the absolute value function is convex, it follows from the Jensen's inequality that
\begin{align}
    &\left| \mathbb{E}_{x \,\sim\, \mathcal{D}}
       \left[\, 
       \mathbb{E}_{y  \,\sim \,\pi_{\theta}}[\,r(x, y)]
       \,\right] - \mathbb{E}_{x \,\sim\, \mathcal{D}}
       \left[\, 
       \mathbb{E}_{y  \,\sim \,\pi }[\,r(x, y)]
       \,\right]  \right| \\
       &\leq  \mathbb{E}_{x \,\sim\, \mathcal{D}}
       \left[\, \left| \sum_{y \in \ccalY} \left(\pi_{\theta} (y \mid x) - \pi (y \mid x) \right) r(x, y) \right| \right].
\end{align}
By triangle inequality, we further obtain
\begin{align}
    &\left| \mathbb{E}_{x \,\sim\, \mathcal{D}}
       \left[\, 
       \mathbb{E}_{y  \,\sim \,\pi_{\theta}}[\,r(x, y)]
       \,\right] - \mathbb{E}_{x \,\sim\, \mathcal{D}}
       \left[\, 
       \mathbb{E}_{y  \,\sim \,\pi}[\,r(x, y)]
       \,\right]  \right| \\
       &\leq  \mathbb{E}_{x \,\sim\, \mathcal{D}}
       \left[\, \sum_{y \in \ccalY} \left| \left(\pi_{\theta} (y \mid x) - \pi (y \mid x) \right) \right| \cdot \left| r(x, y) \right| \right] \\
    &\leq  \mathbb{E}_{x \,\sim\, \mathcal{D}}
       \left[\, \sum_{y \in \ccalY} \left| \left(\pi_{\theta} (y \mid x) - \pi (y \mid x) \right) \right| \cdot B \right],
\end{align}
where the last inequality follows from Assumption~\ref{ass_boundness}. Moreover, Assumption~\ref{ass_parameter_gap} implies that
\begin{align}
    &||\pi_{\theta}(\cdot\,\vert\,x) - \pi(\cdot\,\vert\,x)||_1  \leq \nu. \label{eqn_appen_ass_para_gap1} 
\end{align}

Consequently, it holds by \eqref{eqn_appen_ass_para_gap1} and the definition of $L_1$ norm that
\begin{align}
    &\left| \mathbb{E}_{x \,\sim\, \mathcal{D}}
       \left[\, 
       \mathbb{E}_{y  \,\sim \,\pi_{\theta}}[\,r(x, y)]
       \,\right] - \mathbb{E}_{x \,\sim\, \mathcal{D}}
       \left[\, 
       \mathbb{E}_{y  \,\sim \,\pi}[\,r(x, y)]
       \,\right]  \right|  \leq  \mathbb{E}_{x \,\sim\, \mathcal{D}}
       \left[\, \nu \cdot B \right] = B \nu.
\end{align}

The proof of \eqref{eqn_append_lemma_constraint} is identical to that of \eqref{eqn_append_lemma_reward}, as Assumption~\ref{ass_boundness} indicates both $|r(x, y)| \leq B $ and $|g_i(x, y)| \leq B, \, i = 1,\ldots, m$, for any $(x, y) \in \ccalX \times \ccalY$. This completes the proof of Lemma~\ref{lemma_append_bound_B_nu}.

\end{myproof}

Having introduced Lemma~\ref{lemma_append_bound_B_nu}, we are in the stage of proving Theorem \ref{theorem_primal_dual_gap}.

\textbf{Theorem~\ref{theorem_primal_dual_gap}} (Primal-dual gap).
    Let Assumptions~\ref{ass_boundness}, \ref{ass_parameter_gap} and~\ref{ass_feasibility} hold. Denote by $\tilde{D}(\lambda)$ the dual function of \eqref{eqn_constrained_RL_policy} and define $\lambda_\nu^\star = \argmin_{\lambda\,\geq\;0} \tilde{D}(\lambda) - B\nu\norm{\lambda}_1$. Then, it holds that
    \begin{equation*}
    0
    \; \leq \;
    D^\star  
    -
    P^\star
    \; \leq \;
     \left(\beta+B+B\norm{\lambda_\nu^\star}_1 \right)\nu.
    \end{equation*}
\begin{myproof}
    Notice that the left hand side inequality in \eqref{eqn_primal_dual_gap} follows directly from the standard weak duality theorem~\cite{boyd2004convex}. We then focus on proving the right hand side inequality in \eqref{eqn_primal_dual_gap}. To do so, we rely on the following perturbed problem of \eqref{eqn_constrained_RL_policy}, whose optimal solution is denoted by $\pi_\nu^\star$
\begin{align}\label{eqn_constrained_RL_policy_perturb}
        \tilde{P}^\star (\nu) = \max_{\pi\,\in\,\Pi}
        \; & \;
        \mathbb{E}_{x \,\sim\, \mathcal{D}}
       \left[\, 
       \mathbb{E}_{y  \,\sim \,\pi}[\,r(x, y)\,]
       -
       \beta\,
       D_{\text{KL}}(\pi(\cdot\,\vert\, x) \,\Vert\, \pi_{\text{ref}}(\cdot \, \vert \, x))
       \,\right]\
       \nonumber \\
       \text{s.t.} \; &\;
       \mathbb{E}_{x \,\sim\, \mathcal{D}}
       \left[\, 
       \mathbb{E}_{y  \,\sim \,\pi}[\,g_i(x, y)]\,]
       \,\right] 
       \;\geq\;
       B \nu,\; \, i = 1,\ldots, m
\end{align}
Assumption~\ref{ass_feasibility} indicates the strict feasibility of \eqref{eqn_constrained_RL_policy_perturb} due to the fact that $\pi_\theta \in \Pi$ for any $\theta \in \Theta$. Then strong duality holds for \eqref{eqn_constrained_RL_policy_perturb}~\cite{boyd2004convex}, i.e.,
\begin{align}\label{eqn_strong_dual_perturb}
    \tilde{P}^\star (\nu) = \min_{\lambda \geq 0} \, \max_{\pi \in \Pi} L_\nu(\pi, \lambda),
\end{align}
where $L_\nu(\pi, \lambda)$ represents the Lagrangian of the perturbed problem \eqref{eqn_constrained_RL_policy_perturb}, and can be re-written as
\begin{align}\label{eqn_lagrangian_perturb}
    L_\nu(\pi, \lambda) = L(\pi, \lambda) - B\nu || \lambda ||_1.
\end{align}
Denote by $\lambda_\nu^\star$ the dual optimizer of \eqref{eqn_strong_dual_perturb}. Since $D^\star$ represents the minimum of the dual function $D(\lambda), \forall \lambda \geq 0$, it holds that
\begin{align}
    D^\star \leq D(\lambda_\nu^\star) = \max_{\theta \in \Theta} L(\pi_\theta, \lambda_\nu^\star),
\end{align}
where the last equation follows from the definition of the dual function. Since $\pi_\theta \in \Pi$ for any $\theta \in \Theta$, we can further obtain
\begin{align}
    D^\star \leq \max_{\theta \in \Theta} L(\pi_\theta, \lambda_\nu^\star) \leq \max_{\pi \in \Pi} L(\pi, \lambda_\nu^\star).
\end{align}
Substituting \eqref{eqn_lagrangian_perturb} into the previous inequalities yields
\begin{align}
    D^\star \leq \max_{\pi \in \Pi} L_\nu(\pi, \lambda_\nu^\star) + B\nu || \lambda_\nu^\star ||_1.
\end{align}
By strong duality of \eqref{eqn_constrained_RL_policy_perturb}, the previous inequality reduces to
\begin{align}
    D^\star &\leq \tilde{P}^\star (\nu) + B\nu || \lambda_\nu^\star ||_1 \\
    &=\mathbb{E}_{x \,\sim\, \mathcal{D}}
       \left[\, 
       \mathbb{E}_{y  \,\sim \,\pi_\nu^\star}[\,r(x, y)\,]
       -
       \beta\,
       D_{\text{KL}}(\pi_\nu^\star(\cdot\,\vert\, x) \,\Vert\, \pi_{\text{ref}}(\cdot \, \vert \, x))
       \,\right] + B\nu || \lambda_\nu^\star ||_1, \label{eqn_append_far_away}
\end{align}
where the last equation follows from the definition of the primal optimum.
Adding and subtracting a $P^\star$ in \eqref{eqn_append_far_away} further yields
\begin{align}\label{eqn_append_add_substract_p_star}
    D^\star &\leq P^\star + \mathbb{E}_{x \,\sim\, \mathcal{D}}
       \left[\, 
       \mathbb{E}_{y  \,\sim \,\pi_\nu^\star}[\,r(x, y)\,]
       -
       \beta\,
       D_{\text{KL}}(\pi_\nu^\star(\cdot\,\vert\, x) \,\Vert\, \pi_{\text{ref}}(\cdot \, \vert \, x))
       \,\right] - P^\star + B\nu || \lambda_\nu^\star ||_1.
\end{align}

Notably, it follows from Lemma~\ref{lemma_append_bound_B_nu} that there exists a parameter $\theta_{\lambda_\nu^\star}$ such that 
\begin{align}
    &\left| \mathbb{E}_{x \,\sim\, \mathcal{D}}
       \left[\, 
       \mathbb{E}_{y  \,\sim \,\pi_{\theta_{\lambda_\nu^\star}}}[\,r(x, y)]
       \,\right] - \mathbb{E}_{x \,\sim\, \mathcal{D}}
       \left[\, 
       \mathbb{E}_{y  \,\sim \,\pi_\nu^\star}[\,r(x, y)]
       \,\right] \right| \leq B \nu, \label{eqn_append_lemma_result1} \\
    &\left| \mathbb{E}_{x \,\sim\, \mathcal{D}}
       \left[\, 
       \mathbb{E}_{y  \,\sim \,\pi_{\theta_{\lambda_\nu^\star}}}[\,g_i(x, y)]
       \,\right] - \mathbb{E}_{x \,\sim\, \mathcal{D}}
       \left[\, 
       \mathbb{E}_{y  \,\sim \,\pi_\nu^\star}[\,g_i(x, y)]
       \,\right] \right| \leq B \nu, \; \, i = 1,\ldots, m
\end{align}

Therefore,
\begin{align}
    &\mathbb{E}_{x \,\sim\, \mathcal{D}}
       \left[\, 
       \mathbb{E}_{y  \,\sim \,\pi_{\theta_{\lambda_\nu^\star}}}[\,g_i(x, y)]
       \,\right] \geq \mathbb{E}_{x \,\sim\, \mathcal{D}}
       \left[\, 
       \mathbb{E}_{y  \,\sim \,\pi_\nu^\star}[\,g_i(x, y)]
       \,\right] - B \nu \geq 0, \; \, i = 1,\ldots, m
\end{align}
where the previous inequalities hold by the fact that $\pi_\nu^\star$ is a feasible solution to problem \eqref{eqn_constrained_RL_policy_perturb}. Accordingly, $\theta_{\lambda_\nu^\star}$ is feasible to the original problem \eqref{eqn_constrained_RL_parameterize} whose optimal value is $P^\star$, thus
\begin{align}\label{eqn_append_feasible_to_original}
    P^\star \geq \mathbb{E}_{x \,\sim\, \mathcal{D}}
       \left[\, 
       \mathbb{E}_{y  \,\sim \,\pi_{\theta_{\lambda_\nu^\star}}}[\,r(x, y)\,]
       -
       \beta\,
       D_{\text{KL}}(\pi_{\theta_{\lambda_\nu^\star}}(\cdot\,\vert\, x) \,\Vert\, \pi_{\text{ref}}(\cdot \, \vert \, x))
       \,\right].
\end{align}

Substituting \eqref{eqn_append_feasible_to_original} into \eqref{eqn_append_add_substract_p_star} yields
\begin{align}
    D^\star &\leq P^\star + \mathbb{E}_{x \,\sim\, \mathcal{D}}
       \left[\, 
       \mathbb{E}_{y  \,\sim \,\pi_\nu^\star}[\,r(x, y)\,]
       -
       \beta\,
       D_{\text{KL}}(\pi_\nu^\star(\cdot\,\vert\, x) \,\Vert\, \pi_{\text{ref}}(\cdot \, \vert \, x))
       \,\right] - \nonumber \\
       &\quad\,\, \mathbb{E}_{x \,\sim\, \mathcal{D}}
       \left[\, 
       \mathbb{E}_{y  \,\sim \,\pi_{\theta_{\lambda_\nu^\star}}}[\,r(x, y)\,]
       -
       \beta\,
       D_{\text{KL}}(\pi_{\theta_{\lambda_\nu^\star}}(\cdot\,\vert\, x) \,\Vert\, \pi_{\text{ref}}(\cdot \, \vert \, x))
       \,\right] + B\nu || \lambda_\nu^\star ||_1 \\
       &=P^\star + \mathbb{E}_{x \,\sim\, \mathcal{D}}
       \left[\, 
       \mathbb{E}_{y  \,\sim \,\pi_\nu^\star}[\,r(x, y)\,]  \right]
       - \mathbb{E}_{x \,\sim\, \mathcal{D}}
       \left[\, 
       \mathbb{E}_{y  \,\sim \,\pi_{\theta_{\lambda_\nu^\star}}}[\,r(x, y)\,] \right] + \nonumber \\
       &\quad\, \mathbb{E}_{x \,\sim\, \mathcal{D}}
       \left[\, \beta\,
       D_{\text{KL}}(\pi_{\theta_{\lambda_\nu^\star}}(\cdot\,\vert\, x) \,\Vert\, \pi_{\text{ref}}(\cdot \, \vert \, x))
        -  \beta\,
       D_{\text{KL}}(\pi_\nu^\star(\cdot\,\vert\, x) \,\Vert\, \pi_{\text{ref}}(\cdot \, \vert \, x))
       \,\right] + B\nu || \lambda_\nu^\star ||_1 \\
       &\leq P^\star + \left| \mathbb{E}_{x \,\sim\, \mathcal{D}}
       \left[\, 
       \mathbb{E}_{y  \,\sim \,\pi_\nu^\star}[\,r(x, y)\,]  \right]
       - \mathbb{E}_{x \,\sim\, \mathcal{D}}
       \left[\, 
       \mathbb{E}_{y  \,\sim \,\pi_{\theta_{\lambda_\nu^\star}}}[\,r(x, y)\,] \right] \right| \nonumber \\
       &\quad +  \left| \mathbb{E}_{x \,\sim\, \mathcal{D}}
       \left[\, \beta\,
       D_{\text{KL}}(\pi_{\theta_{\lambda_\nu^\star}}(\cdot\,\vert\, x) \,\Vert\, \pi_{\text{ref}}(\cdot \, \vert \, x))
        -  \beta\,
       D_{\text{KL}}(\pi_\nu^\star(\cdot\,\vert\, x) \,\Vert\, \pi_{\text{ref}}(\cdot \, \vert \, x))
       \,\right] \right| + B\nu || \lambda_\nu^\star ||_1
\end{align}

It then holds by \eqref{eqn_append_lemma_result1} and Assumption~\ref{ass_parameter_gap} that
\begin{align}
    D^\star \leq P^\star + B \nu + \beta \nu + B\nu || \lambda_\nu^\star ||_1.
\end{align}

i.e.,
\begin{align}
    D^\star - P^\star &\leq (\beta + B + B || \lambda_\nu^\star ||_1) \nu.
\end{align}

This completes the proof of Theorem~\ref{theorem_primal_dual_gap}.

\end{myproof}

\section{Example of the Prompt and Response}
\label{append_prompt_response}
The example of the prompt and response in ConstrainedSQL are presented in Figure~\ref{fig_prompt} and Figure~\ref{fig_response}, respectively.

\begin{figure}[!h]
    \centering
    \begin{promptbox}{Prompt}
You are a helpful SQL expert assistant. \\
The assistant first thinks about how to write the SQL query by analyzing the question, database schema and external knowledge, then provides the final SQL query.
The reasoning process and SQL query are enclosed within <think> </think> and <answer> </answer> tags respectively.
The answer must contain the SQL query within \texttt{`\hspace{0.05mm}`\hspace{0.05mm}`sql...`\hspace{0.05mm}`\hspace{0.05mm}`} tags. \\

Database Schema: \{schema\} \\

External Knowledge: \{external\_knowledge\} \\

For example: \\
<think> \\
$\cdots$ \\
</think> \\
<answer> \\
$\cdots$ \\
</answer> \\
\\
Question: \{What is the unabbreviated mailing street address of the school with the highest FRPM count for K-12 students?\} \\
 
    \end{promptbox}
    \caption{The example of the prompt in ConstrainedSQL.}
    \label{fig_prompt}
\end{figure}

\begin{figure}[!h]
    \centering
    \begin{promptbox}{Example Response}

<think>\\
To answer the question, we need to find the school with the highest FRPM count for K-12 students and then retrieve its unabbreviated mailing street address. Here are the steps to generate the SQL query:\\ \\
1. **Identify the relevant table**: The `frpm` table contains the FRPM count information for K-12 students.\\ 
2. **Sort the data by FRPM count**: We need to sort the data in descending order based on the `FRPM Count (K-12)` column to find the school with the highest count.\\
3. **Retrieve the required information**: We need to select the `MailStreet` column from the `schools` table, which contains the unabbreviated mailing street address.\\
4. **Join the tables**: Since the `frpm` table does not contain the mailing street address, we need to join it with the `schools` table using the `CDSCode` column, which is the primary key in both tables.\\
5. **Limit the result**: We only need the school with the highest FRPM count, so we use the `LIMIT 1` clause to restrict the result to one row.\\
</think> \\ \\
<answer>\\
```sql\\
SELECT T2.MailStreet\\
FROM frpm AS T1\\
JOIN schools AS T2 ON T1.CDSCode = T2.CDSCode\\
ORDER BY T1.`FRPM Count (K-12)` DESC\\
LIMIT 1\\
```\\
</answer>

    \end{promptbox}
    \caption{The example of the response in ConstrainedSQL.}
    \label{fig_response}
\end{figure}


 


\end{document}